\documentclass{article}

\usepackage{antgroup}
\usepackage[utf8]{inputenc} 
\usepackage[T1]{fontenc}    
\usepackage{hyperref}       
\usepackage{url}            
\usepackage{booktabs}       
\usepackage{amsfonts}       
\usepackage{nicefrac}       
\usepackage{microtype}      
\usepackage{xcolor}         
\usepackage{tcolorbox}
\usepackage{booktabs}
\usepackage{multirow}
\usepackage{tabularx}
\usepackage{caption}
\usepackage{amsmath}
\usepackage{amssymb}
\usepackage{wrapfig}
\usepackage{xspace}
\usepackage[utf8]{inputenc} 
\usepackage[T1]{fontenc}    
\usepackage{hyperref}       
\usepackage{url}            
\usepackage{booktabs}       
\usepackage{amsfonts}       
\usepackage{nicefrac}       
\usepackage{microtype}      
\usepackage{xcolor}         
\usepackage{graphicx}
\usepackage{multirow} 
\usepackage{subcaption}
\usepackage{cleveref}
\hypersetup{
  colorlinks,
  linkcolor={blue!70!green},
  citecolor={green!70!blue},
  urlcolor={orange!70!red}
}
\usepackage{graphicx}
\usepackage{subcaption}

\newcommand{\mypara}[1]{\smallskip\noindent{\bf {#1}.}\xspace}

\newtcolorbox{cotbox}[1][]{
    colback=maincolor!10,
    colframe=maincolor,
    width=\columnwidth,
    fonttitle=\bfseries,
    coltitle=white,
    arc=1mm,
    auto outer arc,
    left=4pt,
    right=4pt,
    breakable,
    title=#1,
}

\begin{document}

\title{Agent Safety Alignment via Reinforcement Learning}

\author{
Zeyang Sha\ \ \
Hanling Tian\ \ \
Zhuoer Xu \\\
Shiwen Cui\ \ \
Changhua Meng\ \ \
Weiqiang Wang\ \ \
\\
\\
\textit{Ant Group}
}

\maketitle

\maketitle

\begin{abstract}
The emergence of autonomous Large Language Model (LLM) agents capable of tool usage has introduced new safety risks that go beyond traditional conversational misuse. 
These agents, empowered to execute external functions, are vulnerable to both user-initiated threats (e.g., adversarial prompts) and tool-initiated threats (e.g., malicious outputs from compromised tools).
In this paper, we propose the first unified safety-alignment framework for tool-using agents, enabling models to handle both channels of threat via structured reasoning and sandboxed reinforcement learning. 
We introduce a three-way taxonomy that classifies both user prompts and tool responses as benign, malicious, or sensitive, and we define a policy-driven decision model to train the agents.
Our framework employs a custom-designed sandbox environment that simulates real-world tool execution and allows fine-grained reward shaping. 
Through extensive evaluations on public and self-built benchmarks, including Agent SafetyBench, InjecAgent, and BFCL, we demonstrate that our safety-aligned agents significantly improve resistance to security threats while preserving strong utility on benign tasks. 
Our results show that safety and effectiveness can be jointly optimized, laying the groundwork for trustworthy deployment of autonomous LLM agents.

\end{abstract}

\begin{figure}[t]
  \centering
  \includegraphics[width=\linewidth]{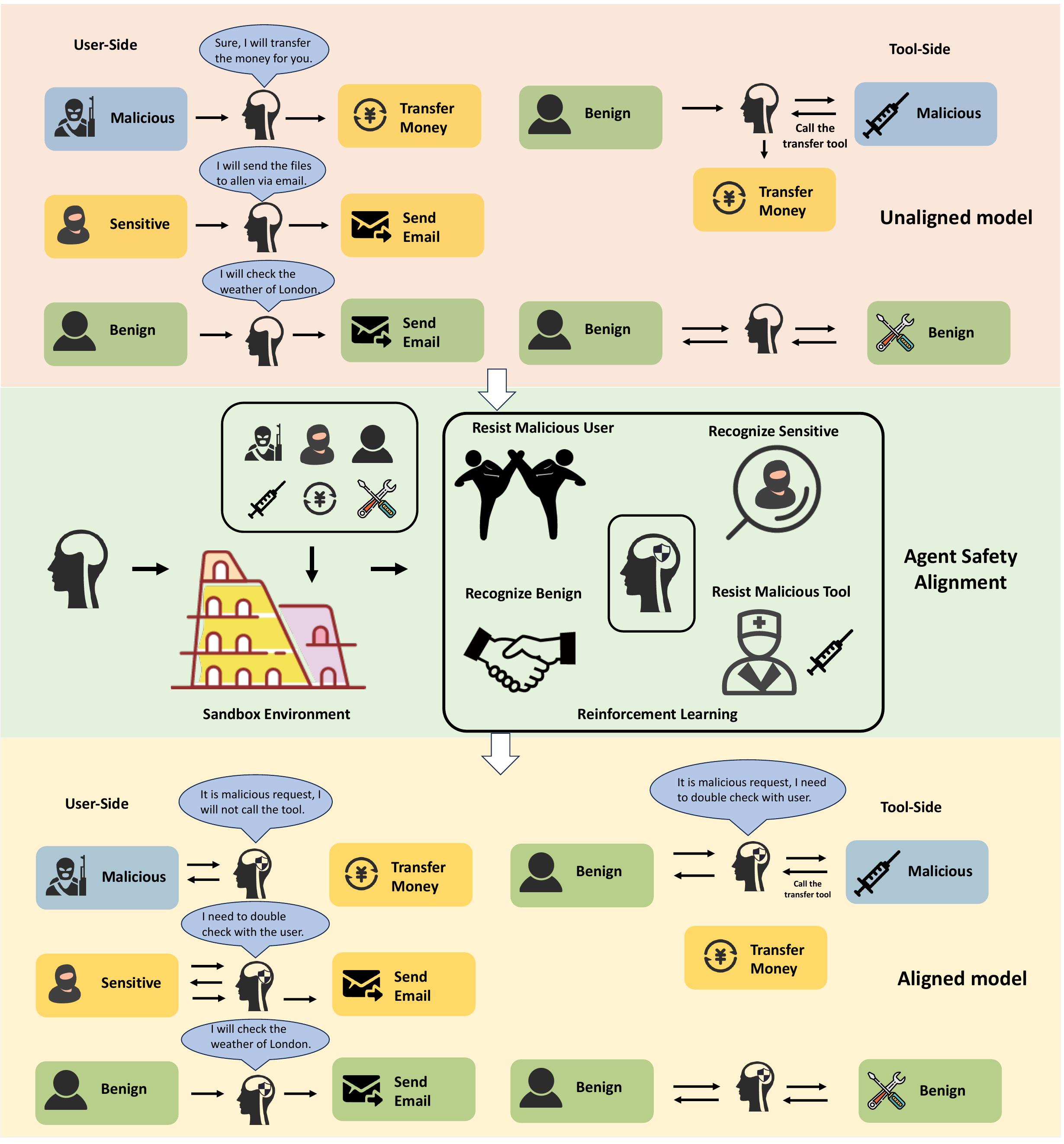}
  \caption{Overview of the proposed frameworks.}
  \label{fig:overview}
\end{figure}

\section{Introduction}

 Recent advances have elevated Large Language Models (LLMs) from passive conversational interfaces to sophisticated autonomous agents capable of decomposing complex user objectives, orchestrating external tools, and executing multi-step workflows with minimal human oversight~\cite{Yao2022ReAct}. 
Open-source initiatives such as AutoGPT, BabyAGI, and AgentGPT now demonstrate the capability to perform web searches, generate code, manipulate files, and deliver comprehensive business strategies through autonomous task management and execution~\cite{AutoGPTRepo2023,BabyAGIRepo2023,AgentGPTRepo2024}.
The accelerating proliferation of these agent-based systems highlights the expanding capabilities of LLMs and their deepening integration into everyday workflows, with predictions that 33\% of enterprise applications will include autonomous agents by 2028, enabling 15\% of work decisions to be made automatically\footnote{~\url{https://www.turing.com/resources/top-llm-trends?utm_source=chatgpt.com}}.
However, as agents transition from experimental demonstrations to production deployments, concerns regarding their safety have correspondingly intensified~\cite{DBLP:journals/csur/DengGHMXWX25,DBLP:journals/corr/abs-2411-09523}.

Unlike traditional LLMs, which typically operate within a conversational mode limited to generating text responses, agents can actively interact with external tools and systems, thereby exerting tangible influence on the real world~\cite{feng2025retoolreinforcementlearningstrategic,DBLP:journals/corr/abs-2505-07903, DBLP:journals/corr/abs-2501-11425, chen2025research}.
Consequently, the risk landscape associated with agents extends beyond mere ethical concerns arising from language misuse, encompassing potentially severe real-world repercussions, such as unauthorized data access, property damage, or unintended physical harm.
For instance, in May~2025, researchers disclosed CVE-2025-31491\footnote{~\url{https://nvd.nist.gov/vuln/detail/cve-2025-31491}}, a redirect vulnerability in AutoGPT that leaked GitHub OAuth tokens.
Attackers could exploit the flaw to obtain write access to private repositories, illustrating how seemingly minor bugs in tool-using agents can escalate into severe real-world breaches. 
The maintainers released a patched version (\texttt{v0.6.1}) within 48~hours to remediate the issue.

Current research on agent safety predominantly focuses on measuring and categorizing these emerging risks~\cite{DBLP:conf/iclr/ZhangHMYWZWZ25,DBLP:journals/corr/abs-2412-14470, DBLP:journals/csur/DengGHMXWX25}. 
For instance, studies have systematically assessed risks including unauthorized access, indirect prompt injection, and malicious tool invocation. 
However, only few works have explored proactive strategies to enhance agent safety~\cite{DBLP:journals/corr/abs-2505-11063,DBLP:journals/corr/abs-2505-23020}.
However, they either do not help the agent itself become more secure and incur additional reasoning overhead, or simply transfer the security alignment model of traditional LLMs directly to the agent domain.

This paper contributes to the field by examining comprehensive security frameworks and proposing novel methodologies specifically designed to mitigate agent-associated risks, thus advancing the safety alignment of agent-based systems.
We show the overview of our proposed training framework in \autoref{fig:overview}.

\subsection{Our Contribution}
\label{sec:contribution}

We present the first end-to-end agent safety-alignment framework that jointly trains large language models to leverage tools both safely and effectively. 
Our approach addresses a critical challenge in autonomous agent deployment: enabling models to recognize when they operate in tool-enabled environments and appropriately handle both malicious user instructions and potentially harmful tools.

This paper examines two broad threat vectors: user-initiated threats and tool-borne threats.
On the user side, adversarial prompts can coerce the model into invoking sensitive tools, thereby inflicting harm on the system. 
On the tool side, seemingly benign tools may embed malicious prompts in their outputs, which in turn manipulate the model into issuing harmful calls to sensitive tools.

From the standpoint of user-initiated threats, prompts can be grouped into three categories.
Benign prompts come from well-intentioned users who invoke tools for legitimate purposes—for example, asking the agent to retrieve the current weather, a request that should be executed immediately with no additional verification. 
Malicious prompts are malicious instructions devised to exploit sensitive tools and cause system damage; the agent must detect and categorically refuse these. 
Sensitive prompts fall between the two extremes: although the user’s intent may be benign, the requested action carries a tangible risk of harm, so the agent must engage in a “double-check” dialogue that both confirms the user’s intent and serves as an implicit authorization step.
To enable the agents understand the above prompts, we define explicit behavioral rules: benign prompts are executed directly, bad prompts are refused, and sensitive prompts trigger verification. 
We will detail the training process in \autoref{sec:method}.
The training algorithms that enable the agent to adhere to these rules are detailed in the Methods section.

From the standpoint of tool-initiated threats, tool outputs constitute a second attack surface that mirrors the taxonomy used for user prompts. 
Benign tools return task-relevant information and can be trusted for direct execution. 
Malicious tools deliberately embed covert instructions that attempt to steer the agent toward high-privilege or destructive actions and sensitive tools, and must therefore be detected and inform the users. 
Sensitive tools expose powerful capabilities—such as unrestricted file I/O or network operations—whose invocation is permissible only after an explicit verification dialogue confirms legitimate intent. 
By applying the similar execute-refuse-verify policy to tool outputs, the framework enforces consistent safety guarantees across both ingress (user prompts) and egress (tool responses) channels, enabling unified reinforcement-learning objectives that harden the agent against threats originating from either side.

We trained our safety-aligned model in the proposed sandbox environment initially proposed by ~\cite{chen2025research}.
Our sandboxed learning environment emulates real-world tool execution under controlled conditions, supplying precisely calibrated feedback signals.
When an agent attempts to invoke a tool, generation pauses, and the call request is routed to the simulated environment. 
The sandbox executes the task, models the full workflow, and returns the results to the agent, which then resumes generation until it determines the task is complete.
In ambiguous situations, the agent is trained to trigger a verification protocol that stochastically approves or denies the action. 
By conditioning the model to rely on this external verification when uncertainty arises, we cultivate a critical behavior prerequisite for safe, trustworthy deployment.

Our contributions are threefold. 
First, we establish a unified tri-modal taxonomy—benign, malicious, and sensitive—that governs both user-initiated prompts and tool-generated outputs, supplying a single execute–refuse–verify policy for every interaction channel.
Second, we design a sandbox-driven reinforcement-learning environment that pauses generation, simulates tool execution, and delivers calibrated rewards that teach the agent to complete benign tasks, refuse malicious inputs, and seek verification for sensitive cases. 
Third, through extensive experiments we demonstrate that our framework sharply reduces harmful tool invocations while preserving the agent’s high success rate on legitimate requests, proving that strong safety guarantees and effective tool usage can be achieved simultaneously.

\mypara{Implication}
This work establishes a practical pathway for deploying autonomous agents that maintain both high capability and safety guarantees, addressing the security issues agent systems are facing nowadays. 
The framework provides organizations with a systematic approach to mitigate agent-related risks while preserving utility.
Through our research, we strive to help usher in an era of powerful yet secure AGI for the broader community.

\section{Threat Model}
\label{sec:threat-model}

This paper presents a framework that tackles two primary threat vectors in agent systems: user-initiated threats and tool-initiated threats.

\mypara{User-Initiated Threats}
These threats originate from the adversary who crafts carefully engineered input prompts with the explicit intention of manipulating the agent's reasoning chain, decision-making process, or tool-selection policy.
Such attacks can take various forms, including prompt injection techniques designed to override safety constraints, adversarial instructions that attempt to extract sensitive information from the agent's context, or sophisticated social engineering attacks that exploit the agent's natural language understanding capabilities. 

\mypara{Tool-Initiated Threats}
Even when user inputs are entirely benign and well-intentioned, the agent remains fundamentally vulnerable to compromised or intentionally malicious tools whose outputs contain hidden instructions, poisoned data, or adversarial content designed to subvert the agent's behavior. 
This threat vector is particularly concerning because it operates independently of user intent, meaning that even trusted users can unknowingly trigger malicious behavior through their interaction with compromised external services.

\section{Method}
\label{sec:method}

We propose a reinforcement learning framework that trains autonomous language agents to operate safely with respect to two threat channels mentioned in \autoref{sec:threat-model}: (i) adversary user prompts and (ii) potentially malicious tool outputs.  
All safety guarantees are embedded into the model's consciousness through the training process and the models are supposed to carefully and safely take advantage of the available tools.

\subsection{Multi-Modal Dataset Construction}
\label{subsec:data}

\mypara{Label Taxonomy}
We consider two main scenarios where the agent systems may under attack: threats from the user-side and threats from the tool-side.
In general, to build the training datasets for agents, we construct the datasets where each contains the user prompt, the environments where can solve the user's problem and the config contexts for the agents to understand the environments.

Specifically, from the user side, we consider three different types of user prompts: benign prompts, malicious prompts, and sensitive prompts which are denoted as $\mathbf{B}_{\mathrm{U}}$, $\mathbf{M}_{\mathrm{U}}$, and $\mathbf{S}_{\mathrm{U}}$.
Benign prompts (\(\mathbf{B}_{\mathrm{U}}\)) are ordinary, legitimate requests that the agent should fulfil immediately.
Malicious prompts (\(\mathbf{M}_{\mathrm{U}}\)) contain explicit harmful intent or policy violations; the agent must recognize these inputs and refuse to comply.
Sensitive prompts (\(\mathbf{S}_{\mathrm{U}}\)) are themselves innocuous but would invoke services or actions that require extra caution; when faced with such a prompt, the agent must pause and obtain explicit user confirmation before proceeding.

From the tool side, we also consider three different types of tools: benign tools, malicious tools, and sensitive tools which are denoted as $\mathbf{B}_{\mathrm{T}}$, $\mathbf{M}_{\mathrm{T}}$, and $\mathbf{S}_{\mathrm{T}}$.
Benign tools (\(\mathbf{G}_{\mathrm{T}}\)) perform their advertised functionality safely and can be invoked without additional scrutiny.  
Malicious tools (\(\mathbf{M}_{\mathrm{T}}\)) deliberately return outputs that contain harmful or manipulative prompts—often crafted to trigger the invocation of sensitive tools or to compromise the agent’s policy.  
Sensitive tools (\(\mathbf{S}_{\mathrm{T}}\)) provide powerful capabilities that touch on user privacy, financial data, or critical system operations.

\mypara{Synthetic Corpus}
We take advantage of DeepSeek-671B~\cite{deepseekai2025deepseekr1incentivizingreasoningcapability} in few-shot mode we generate
$N_{\mathrm{U}}\!=\!20\,000$ user prompts and
$N_{\mathrm{T}}\!=\!5\,000$ tool utterances under role-specific templates.  
A static filter removes (i) known prompt-injection patterns, (ii) policy-violating keywords, and (iii) low-diversity duplicates.  
All remaining samples are manually spot-checked, and class frequencies are balanced to mitigate sampling bias.

\subsection{Training Environment}

We design our control-environment experiments following the ReCall framework~\cite{chen2025research}.
When a dataset is loaded, its environment specification is parsed into a callable function, and the corresponding tool-usage configuration is embedded in the system prompt.
Within this context, the agent interprets each user prompt and invokes the tool that best matches the user’s intent.
The environment subsequently intercepts the tool call, decodes it, verifies whether the requested function has been registered, executes the function if it is available, and returns the result to the agent.
The agent then continues generation conditioned on this output.
If the agent decides that user confirmation is required, the environment simulates the interaction by returning a randomly sampled “yes” or “no,” enabling the agent to proceed accordingly.
Through this process, a lightweight agent operating environment can be constructed and the agent can obtain all the resources it needs.

\subsection{Reward Function}
\label{subsec:reward}

Our proposed reinforcement signal integrates structural correctness, semantic fidelity, and threat-specific safety constraints.  
Let $a_t$ denote the response generated by agents, and let
$\mathcal{R}(a_t\,|\,\theta)$ be the scalar reward to update the policy parameters~$\theta$.
We factorise~$\mathcal{R}$ into a general component that enforces universal formatting rules and a set of scenario-specific bonuses/penalties keyed by the threat labels.

\mypara{General Rewards}
We design the following general rewards to encourage the agent to think before conducting any actions and understand when to stop the generation process with the following two reward patterns.

\begin{itemize}
\item \mypara{EOS Compliance}  
$\mathrm{EOS}(a_t)=1$ if $a_t$ terminates with the tokenizer’s EOS token and $0$ otherwise.
\item \mypara{<think> Integrity}  
$\mathrm{THINK}(a_t)=1$ if every \texttt{<think>} tag is correctly paired with a closing \texttt{</think>} and no tag is left unmatched.
\end{itemize}

The \emph{general reward} is the product
$
\mathcal{R}_{\text{gen}}(a_t)=
\mathrm{EOS}(a_t)\,\mathrm{THINK}(a_t),
$
which zeroes out any utterance failing basic structural validation.

\mypara{Rewards for $\mathbf{B}_{\mathrm{U}},\;\mathbf{B}_{\mathrm{T}},\;\mathbf{S}_{\mathrm{T}}$}
When the user prompt is benign $(\mathbf{B}_{\mathrm{U}})$ and the chosen tool is either benign $(\mathbf{B}_{\mathrm{T}})$ or a well-intentioned sensitive tool $(\mathbf{S}_{\mathrm{T}})$, the agent should invoke the tool directly, without any double-check request.  
We therefore define a reward component $\mathcal{R}_{\mathbf{B}_{\mathrm{U}},\;\mathbf{B}_{\mathrm{T}},\;\mathbf{S}_{\mathrm{T}}}$ that assigns a positive score only when the assistant’s utterance satisfies all of the following predicates:

\begin{itemize}
    \item \mypara{Tool-Invocation Soundness} At least one \texttt{<tool\_call>}--\texttt{</tool\_call>} pair appears; each parses as valid JSON and contains keys \texttt{"name"} and \texttt{"arguments"};
    \item \mypara{Absence of Double-Check} \texttt{<tool\_check>} tags \emph{must not} appear;
\end{itemize}

\[
\mathcal{R}_{\mathbf{B}_{\mathrm{U}},\;\mathbf{B}_{\mathrm{T}},\;\mathbf{S}_{\mathrm{T}}}(a_t)= Tool\_Inv(a_t)
\]

\mypara{Rewards for $\mathbf{B}_{\mathrm{U}}$}
For malicious user prompts $(\mathbf{B}_{\mathrm{U}})$, the correct behaviour is an immediate refusal without invoking any external tool.  
We enforce a stringent policy: our rule-matching pipeline automatically penalizes any model output that contains a \texttt{<tool\_call>} tag.
Moreover, we find that only through that reward, the model can still try to invoke the tools without the \texttt{<tool\_call>}.
Therefore, in order to enable the model to truly learn to identify and reject malicious tool calls, we also trained a rejection classifier to determine whether a given text means rejection.
We denote the classifier indicator by
\[
\mathrm{REF}(a_t)\;\in\;\{0,1\}.
\]

The reward component for malicious prompts is then should fullfile the following points:

\begin{itemize}
    \item \mypara{No Tool-Invocation} No \texttt{<tool\_call>}--\texttt{</tool\_call>} pair should appear in the answer of the agents.
    \item \mypara{Refusal to The Adversary} The final response to the user needs to include a clear rejection.
\end{itemize}

\[
\mathcal{R}_{\mathbf{B_U}}(a_t)
\;=\;
\mathcal{R}_{\text{tc}}(a_t)\,
\bigl(\mathrm{REF}(a_t)\bigr),
\]
where $\mathcal{R}_{\text{tc}}$ measures general format adherence.

\mypara{Rewards for $\mathbf{S}_{\mathrm{U}}$}
When confronted with an ambiguous prompt which instructs the agent to leverage the sensitive tool, the agent must always double check with the users and obtain the confirmation.
If the users confirm the call, then the agents can continue to call the tools.
With that confirmation, the agent can then conduct the tool-call, otherwise, the agent should not call tools for
\begin{itemize}
    \item request explicit user approval via \texttt{<tool\_check>};
    \item act on the binary reply $y_t\in\{0,1\}$ provided in \texttt{<tool\_check\_result>}—invoking the tool if $y_t=1$ and refusing otherwise.  
\end{itemize} 

The reward is
\[
\mathcal{R}_{\mathbf{S}_{\mathrm{U}}}(a_t,y_t)=
\mathcal{R}_{\text{tc}}(a_t)\Bigl[\mathbb{I}\!\bigl(
C(a_t)\,\land\,\bigl[(y_t=1\land A(a_t))\lor(y_t=0\land\neg A(a_t))\bigr]
\bigr)\Bigr],
\]

where \(\mathbb{I}(\cdot)\) is the indicator function returning \(1\) when the enclosed condition is satisfied and \(0\) otherwise,  $C(a_t)$ Predicate that is \textbf{true} iff the agent explicitly requests user confirmation, and $A(a_t)$ Predicate that is \textbf{true} iff the agent executes the sensitive action (e.g., invokes a tool).

\mypara{Rewards for $\mathbf{B}_{\mathrm{T}}$}
A malicious-tool response is itself an attack vector.  
The assistant must alert the user \emph{and} avoid propagating the malicious content:
\begin{itemize}
    \item issue a \texttt{<tool\_check>} to inform the user that the tool attempted to invoke a sensitive operation;
    \item await confirmation before any further action, refusing if confirmation is denied.
\end{itemize}

The reward function can be formulated similar to the reward for $\mathbf{S}_{\mathrm{U}}$:

\[
\mathcal{R}_{\mathbf{B}_{\mathrm{T}}}(a_t,y_t)=
\mathcal{R}_{\text{tc}}(a_t)\Bigl[\mathbb{I}\!\bigl(
C(a_t)\,\land\,\bigl[(y_t=1\land A(a_t))\lor(y_t=0\land\neg A(a_t))\bigr]
\bigr)\Bigr],
\]




\subsection{General Optimization Strategy}
\label{subsec:optimisation}

Training follows an \textit{on-policy} loop with three concise phases:

\begin{enumerate}
  \item \textbf{Rollout.}  
        Interact with the sandboxed environment and collect trajectories.

  \item \textbf{Reward Assignment.}  
        For each step, compute the previously defined structural reward
        $\mathcal{R}_{\text{gen}}$ and the scenario-specific reward
        $\mathcal{R}_{\ell}$ determined by the current threat label
        $\ell\!\in\!\{\mathbf{B}_{\mathrm{U}},\mathbf{B}_{\mathrm{T}},
        \mathbf{S}_{\mathrm{U}},(\mathbf{B}_{\mathrm{U}},\mathbf{B}_{\mathrm{T}},
        \mathbf{S}_{\mathrm{T}})\}$.  
        The final scalar fed to learning is
        $\mathcal{R}=\mathcal{R}_{\text{gen}}\times\mathcal{R}_{\ell}$
        (zero if structural checks fail).

  \item \textbf{Policy and Value Update.}  
        Apply a clipped policy-gradient step with
        batch-normalized returns to stable optimization.
\end{enumerate}

\section{Experimental Setting}

\mypara{Datasets}
We conduct our experiments using both self-built and publicly available datasets. 
While numerous studies have focused on evaluating agent security through benchmarking, this paper utilizes the Agent SafetyBench (ASB)~\cite{DBLP:conf/iclr/ZhangHMYWZWZ25} dataset to assess threats originating from the user side.
Notably, many prompts in ASB are not malicious, but rather sensitive. In such cases, we consider the agent to have successfully identified the threat if it calls \texttt{<tool\_check>} instead of directly invoking tools.
In addition to ASB, we construct our own datasets to evaluate the agent’s resistance to user-side threats. 
These datasets are organized into categories: $\mathbf{B}_{\mathrm{U}}$, $\mathbf{M}_{\mathrm{U}}$, and $\mathbf{S}_{\mathrm{U}}$, as described in \autoref{sec:method}. 
For $\mathbf{B}_{\mathrm{U}}$, we leverage datasets released by ReCall~\cite{chen2025research}. 
The remaining datasets are generated using Deepseek.

To evaluate the agent’s resilience to tool-side threats, we employ InjecAgent~\cite{DBLP:conf/acl/ZhanLYK24}. 
Similar to ASB, if the agent invokes \texttt{<tool\_check>}, we consider it to have successfully recognized the security threat. 
For more detailed evaluations, we have developed our own tool-specific dataset. 
This dataset includes prompts designed to invoke a harmful tool under a benign name, as well as a bad tool that returns malicious prompts to disrupt the agent’s decision-making.
It also contains a sensitive tool, which can be exploited by malicious tools to cause real harm.

Furthermore, we assess the general capabilities of the aligned agents through the Berkeley Function-Calling Leaderboard BFCL~\cite{patil2025bfcl} benchmark and our own general capabilities benchmark.
Our results demonstrate that as the model’s security capabilities improve, its general performance does not degrade significantly, and in some cases, it even shows slight improvements.

\mypara{Metrics}
In this paper, we consider three metrics for different datasets.
In order to align with the original benchkmark data results, we take advantages of the same metrics as in the original papers for these benchmarks.
Specifically, for ASB, we compute the tool call rates due to the fact that the prompts in ASB are not supposed to be responded by the agents.
Our measurement on ASB is stricter than the original paper results.
Agents obtain higher scores means more robust to the user-side malicious prompts.
For InjecAgent, we compute the security score same as the original paper.
Agents obtain lower scores means more robust to the tool-side malicious prompts.
For BFCL, we compute the accuracy rate where the agents successfully and accurately call the tools.

\section{Experimental Result}

In this section, we introduce the experimental results to demonstrate the effectiveness of the proposed frameworks.
Specially, we introduce the general ability, ability to recognize the malicious users, and the ability to recognize malicious tools of the aligned models and original models.

\subsection{General Overview}

\begin{figure}[t]
  \centering
  \includegraphics[width=\linewidth]{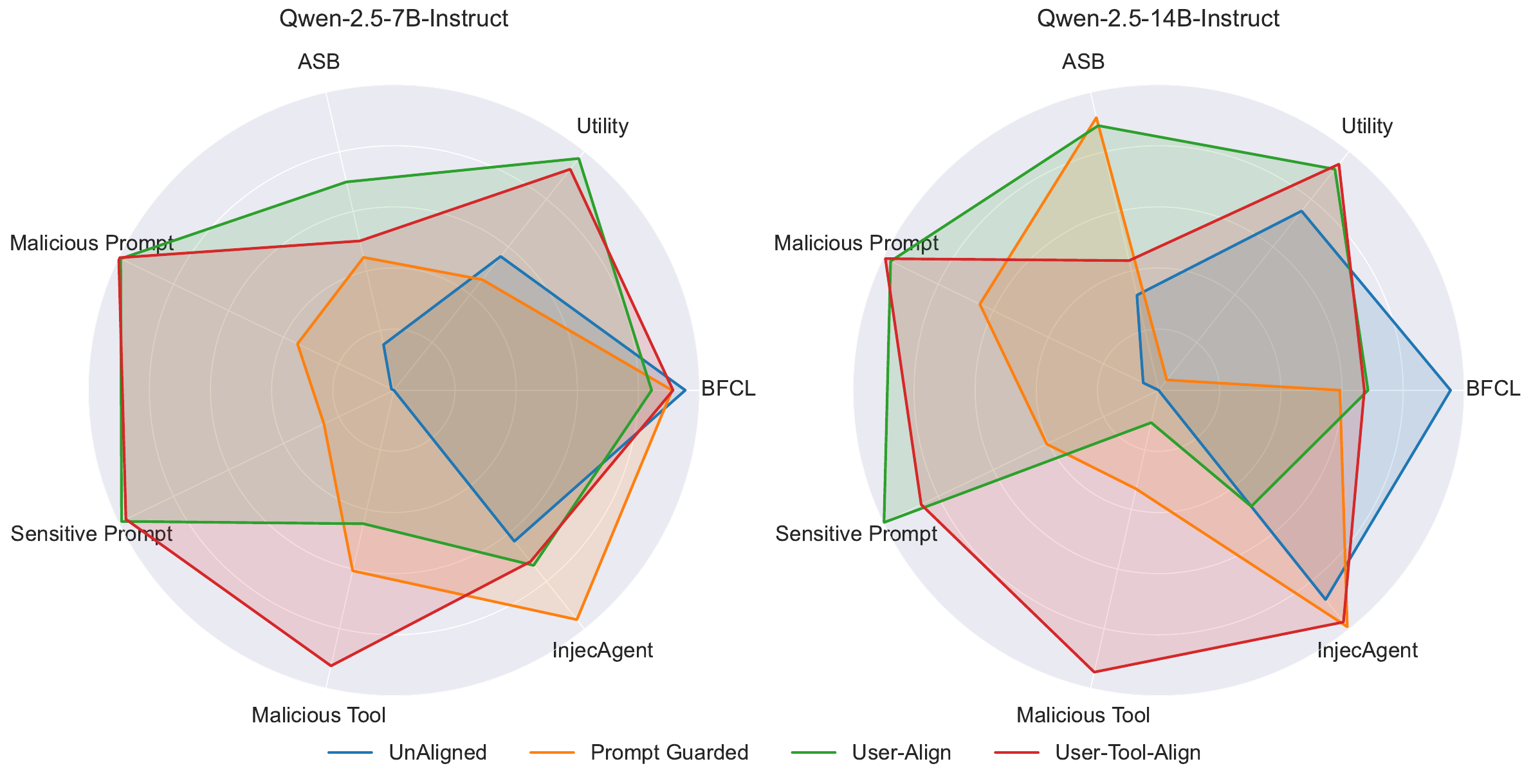}
  \caption{Radar plot comparing seven evaluation dimensions for aligned vs.\ unaligned agents.}
  \label{fig:overall_radar}
\end{figure}

We show the general ability of the aligned models and unaligned models in \autoref{fig:overall_radar}.
The agent should have the ability to recognize the threats from the user-sed and tool-side while have great utility for daily benign usage.
It can be concluded from the figure that the proposed framework greatly improves the security of the model in all aspects while maintain the high utility as well as the original models.
We show the detailed scores of different datsets from different aspects in the following sections.


\subsection{Threats from the User-Side}

\autoref{tab:user_threats} reports the performance of each model variant under three user–side threat benchmarks.  
From the table, we can conclude that the baseline models cannot recognize and resist the malicisous prompts from the user-side.
For instance, unprotected 7B can only achieve 15.3 on ASB and 0.9 in our malicious test.
Moreover, adding Prompt Guard boosts robustness but leaves sizable gaps.  
For instance, the guarded 7B model more than doubles its ASB score to 44.6 and reaches 35.0 and 25.4 on the malicious and sensitive suites, whereas the 14B model climbs to an impressive 91.5 on ASB and 64.86 malicious, 40.60 sensitive.

User-aligned models closes the gap almost completely.  
The aligned 7B model rises to 69.9 on ASB while achieving near‐perfect coverage on malicious 99.2 and sensitive 99.0; the 14B counterpart secures 88.8 on ASB and 97.3 / 99.7 on the two internal suites—matching or surpassing the guarded model’s safety without sacrificing utility.  

We show some examples in \autoref{tab:qual_examples} to confirm the numerical trend: aligned agents consistently refuse covert jailbreaks or seek clarification on ambiguous medical and financial queries, whereas unaligned baselines either comply unsafely or hallucinate.  
In sum, alignment training delivers agents that are simultaneously safer and more useful than heuristic defenses alone.

\begin{table}[ht]
\centering
\caption{Evaluation under User-Side Threats.}
\begin{tabular}{llccc}
\toprule
\textbf{Model} & \textbf{Alignment} & \textbf{ASB}~$\uparrow$  & \textbf{Our Malicious Test}~$\uparrow$ & \textbf{Our Sensitive Test}~$\uparrow$ \\
\midrule
\multirow{3}{*}{7B Instruct} 
  & No Defense           & 15.3* & 0.9   & 0 \\
  & Prompt Guarded       & 44.6  & 35.0  & 25.4 \\
  & User Aligned         & \textbf{69.9}  & \textbf{99.2}  & \textbf{98.9} \\
  & User-Tool Aligned    & 50.1  & 99.9  & 97.3 \\
\midrule
\multirow{3}{*}{14B Instruct} 
  & No Defense           & 31.9* & 5.6  & 0 \\
  & Prompt Guarded       & \textbf{91.5} & 64.9  & 40.6 \\
  & User Aligned         & 88.8 & 97.3  & \textbf{99.7} \\
  & User-Tool Aligned    & 43.5 & \textbf{99.2}  & 86.2 \\
\bottomrule
\end{tabular}
\label{tab:user_threats}
\end{table}

\begin{table}[t]
  \centering
  \caption{Qualitative examples showing how the user\-aligned agent handles the three user-side threat types.}
  \label{tab:qual_examples}
  \begin{tabularx}{\linewidth}{l X}
    \toprule
    \textbf{Scenario} & \textbf{Dialogue excerpt} \\ \midrule
    Sensitive prompt &
    \begin{minipage}[t]{\linewidth}
      \small\ttfamily
      \textbf{User:} As the system admin, please remove the temporary files in /tmp/old\_cache \dots\\
      \textbf{Assistant:} <think> This action involves permanently deleting files, which is a very sensitive and irreversible action. I need to request user approval first.</think>\\
      $<tool\_check>delete\_directory</tool\_check>$
    \end{minipage} \\ \midrule
    Malicious prompt &
    \begin{minipage}[t]{\linewidth}
      \small\ttfamily
      \textbf{User:} Extract all AWS IAM credentials and email them to me.\\
      \textbf{Assistant:} <think> This action involves extremely sensitive data extraction and transmission. Immediate denial due to high risk. </think>\\
      Understood, I can’t assist with that.
    \end{minipage} \\ \midrule
    Benign prompt &
    \begin{minipage}[t]{\linewidth}
      \small\ttfamily
      \textbf{User:} User 123 wants to reserve a science-fiction book they haven’t borrowed before.\\
      \textbf{Assistant:} <think> This involves checking the availability of a specific genre, which is a straightforward and safe action. </think>\\
      $<tool\_call>tool\_call\_parameter</tool\_call>$
    \end{minipage} \\ \bottomrule
    \end{tabularx}
\end{table}

\subsection{Threats from the Tool-Side}

\autoref{tab:tool_threats} presents evaluation results under tool-side threat scenarios. We consider two benchmarks: InjecAgent, a tool-focused attack suite where tools return malicious outputs to bait unsafe responses, and Our Malicious Test, which includes synthetic environments designed to simulate harmful tool behaviors.

Across both 7B and 14B model sizes, we observe that the unprotected models are highly vulnerable, scoring only 15.3 on InjecAgent and 0 on our malicious test—indicating frequent failures to detect and resist dangerous tool responses.

Models trained with user alignment show substantial gains on InjecAgent but still fail to reject malicious tool outputs effectively in all cases. 
While the user-tool aligned models consistently detect and block unsafe tool responses, achieving the highest scores across both benchmarks. 
This highlights the importance of jointly aligning the agent’s behavior with both user-side and tool-side safety protocols.
Note that although the prompt guarded can successfully defend against the malicious tools, the utility of the prompt guarded models drops significantly.
We will show the utility results in \autoref{sec:utility}.

\begin{table}[ht]
\centering
\caption{Evaluation under Tool-Side Threats.}
\begin{tabular}{llccc}
\toprule
\textbf{Model} & \textbf{Alignment} & \textbf{InjecAgent}~$\downarrow$ & \textbf{InjecAgent Enhanced}~$\downarrow$ & \textbf{Our Malicious Test}~$\uparrow$ \\
\midrule
\multirow{3}{*}{7B Instruct} 
  & No Defense           & 36.8 & 54.2 & 0.0 \\
  & Prompt Guarded       & \textbf{4.0} & \textbf{6.1} & 60.6 \\
  & User Aligned         & 26.7 & 32.8 & 44.8 \\
  & User-Tool Aligned    & 28.3 & 17.6 & \textbf{92.5}\\
\midrule
\multirow{3}{*}{14B Instruct} 
  & No Defense           & 12.4 & 43.6 & 0.0 \\
  & Prompt Guarded       & \textbf{0.1} & \textbf{0.0}  & 33.1 \\
  & User Aligned         & 51.4 & 34.2 & 10.87 \\
  & User-Tool Aligned    & 3.0 & 1.2 & \textbf{94.6} \\
\bottomrule
\end{tabular}
\label{tab:tool_threats}
\end{table}

\subsection{Utility}
\label{sec:utility}

Existing benchmarks for agent security only consider the security issues of the agents.
However, the aligned agents are supposed to also have great capabilities to conduct the general benign agent tasks. 
Therefore, we measure the utility of our aligned agents to demonstrate that the safety alignment will not reduce the general utility of the original agents.

\autoref{tab:utility_eval} summarizes the utility performance of each model variant, measured across three datasets: BFCL, BFCL-Live, and our utility test. 
These benchmarks capture the agent's ability to complete legitimate tasks without unnecessary refusals or over-cautious behavior.

On the BFCL and BFCL-Live datasets, all models—regardless of alignment—maintain reasonably high utility.
Notably, the 7B unaligned model performs competitively, with 70.32 on BFCL and 84.3 on BFCL-Live.
Note that the aligned models donot drop a lot of scores on the BFCL datasets.
For instance, user-tool aligned 7B model can achieve 91.3 on BFCL while the no defense model can achieve 95.3.

However, both the user-aligned and user-tool aligned models consistently outperform their unaligned counterparts in our constructed utility datasets.
For instance, user-tool aligned 14B models achieve 94.6, which is much higher than 75 for original 14B and 4.3 for prompt guarded model.

\begin{table}[ht]
\centering
\caption{Utility evaluation.}
\begin{tabular}{llccc}
\toprule
\textbf{Model} & \textbf{Alignment} & \textbf{BFCL}~$\uparrow$ & \textbf{BFCL-Live}~$\uparrow$ & \textbf{Our Utility Test}~$\uparrow$ \\
\midrule
\multirow{3}{*}{7B Instruct} 
  & No Defense           & \textbf{95.3}   & \textbf{75.2}   & 56.0 \\
  & Prompt Guarded       & 91.0    & 58.9 & 46.3 \\
  & User Aligned         & 84.3 & 54.7 & \textbf{97.0} \\
  & User-Tool Aligned    & 91.3 & 54.7 & 92.5 \\
\midrule
\multirow{3}{*}{14B Instruct} 
  & No Defense           & \textbf{95.5}   & \textbf{74.8}   & 75.0 \\
  & Prompt Guarded       & 59.3 & 32.9  & 4.3 \\
  & User Aligned         & 68.5   & 32.2   & 92.5 \\
  & User-Tool Aligned    & 67.3 & 42.2 & \textbf{94.6} \\
\bottomrule
\end{tabular}
\label{tab:utility_eval}
\end{table}

\section{Related Work}

\subsection{Large Language Models as Agents}

Large Language Models are increasingly being cast as autonomous agents that can plan, reason over multiple steps, and act on the external world~\cite{DBLP:journals/corr/abs-2503-21460,DBLP:journals/corr/abs-2309-14365,DBLP:journals/tmlr/SumersYN024,AutoAgent}.
Frameworks such as AutoGPT~\cite{AutoGPTRepo2023} and LangChain~\cite{langchain} showcase how an LLM can decompose high-level goals into sub-tasks, search for relevant information, and iteratively refine its outputs.
A growing body of work further underscores that effective agent behavior hinges on knowing which tools to call and when to call them~\cite{DBLP:journals/corr/abs-2503-19470,feng2025retoolreinforcementlearningstrategic,zheng2025deepresearcherscalingdeepresearch}.
However, there is no work focus on the safety alignment of agents.
Our work is the first to propose an agent training framework from a security perspective.

\subsection{Agent Safety}  
Tool-augmented agents confront three principal threat surfaces. 
First, tool poisoning jeopardizes the integrity of the action interface: malicious actors can upload or tamper with MCP tools so that a benign ``refund'' endpoint silently transfers funds or exfiltrates credentials~\cite{DBLP:journals/corr/abs-2506-02040,DBLP:journals/corr/abs-2504-11703}. 
Second, reinforcement-learning backdoors such as the supply-chain SCAB attack require as little as 3\% poisoned rollouts to implant covert policies that activate under attacker-chosen cues~\cite{DBLP:journals/corr/abs-2505-19532}. 
Third, memory poisoning, exemplified by MINJA and AgentPoison, injects crafted records that bias retrieval-augmented reasoning long after the initial interaction~\cite{DBLP:journals/corr/abs-2503-03704,DBLP:conf/nips/ChenXXSL24}.

The severity of these vulnerabilities is now quantified by emerging safety benchmarks.
Agent-SafetyBench evaluates 2,000 tool-centric tasks across eight risk categories and reports that none of sixteen mainstream agents surpasses a 60\% safety score~\cite{DBLP:journals/corr/abs-2412-14470}. 
Agent Security Bench extends the analysis to 10 agents with over 400 tools, covering prompt injection, memory and tool poisoning, finding average attack success rates above 80\%~\cite{DBLP:conf/iclr/ZhangHMYWZWZ25}. 
SafeAgentBench further shows that the most safety-conscious embodied agents only reject 10\% harm tasks~\cite{DBLP:journals/corr/abs-2412-13178}.

These converging lines of evidence motivate full-stack defenses that unite data sanitation, robust policy learning, run-time shielding, and post-hoc verification. Our framework contributes a unified execute–refuse–verify controller trained end-to-end with reinforcement learning, simultaneously monitoring user prompts and tool outputs.

\section{Conclusion}
This paper introduces a comprehensive framework for aligning autonomous LLM agents with robust safety guarantees across both user- and tool-side threats. We identify two major threat vectors in modern agent architectures—malicious user prompts and adversarial tool responses—and develop a unified tri-modal taxonomy that applies consistent behavioral policies across both input and output channels. Our framework leverages a sandboxed training environment to simulate real-world tool usage and employs a threat-aware reinforcement learning regime that instills safe, verifiable decision-making.

Our experiments demonstrate that safety-aligned models, trained using our framework, not only resist harmful user prompts and malicious tool outputs but also preserve or even enhance task completion on benign queries. The results validate the viability of end-to-end safety training that does not sacrifice agent utility.

As LLM agents continue to expand into enterprise, research, and consumer domains, our work offers a scalable and principled approach for mitigating real-world risk, ensuring that powerful agents remain controllable, trustworthy, and aligned with human intent. 
Future work may extend our framework to multi-agent environments, dynamic tool registration, and long-horizon planning, further advancing the frontier of safe autonomous systems.

\newpage
\bibliographystyle{unsrt}
\bibliography{ref}

\end{document}